# Towards a Method for Synthetic Generation of Persons with Aphasia Transcripts

Jason M. Pittman[1], Anton Phillips Jr.[2], Yesenia Medina-Santos[2], Brielle C. Stark[2]
[1]University of Maryland Global Campus
[2]Indiana University Bloomington, Department of Speech, Language and Hearing Sciences

## ABSTRACT

In aphasia research, Speech-Language Pathologists (SLPs) devote extensive time to manually coding speech samples using Correct Information Units (CIUs), a measure of how informative an individual sample of speech is. Developing automated systems to recognize aphasic language is limited by data scarcity. For example, only about 600 transcripts are available in AphasiaBank yet billions of tokens are used to train large language models (LLMs). In the broader field of machine learning (ML), researchers increasingly turn to synthetic data when such are sparse. Therefore, this study constructs and validates two methods to generate synthetic transcripts of the AphasiaBank Cat Rescue picture description task. One method leverages a procedural programming approach while the second uses Mistral 7b Instruct and Llama 3.1 8b Instruct LLMs. The methods generate transcripts across four severity levels (Mild, Moderate, Severe, Very Severe) through word dropping, filler insertion, and paraphasia substitution. Overall, we found, compared to human-elicited transcripts, Mistral 7b Instruct best captures key aspects of linguistic degradation observed in aphasia, showing realistic directional changes in NDW, word count, and word length amongst the synthetic generation methods. Based on the results, future work should plan to create a larger dataset, fine-tune models for better aphasic representation, and have SLPs assess the realism and usefulness of the synthetic transcripts.

## Introduction

Per Nicholas and Brookshire (1993), coding Correct Information Units (CIUs) involves transcribing a connected speech sample verbatim, counting all intelligible words, and then identifying each word that is intelligible, accurate, relevant, and informative about the topic as a CIU—excluding fillers, repetitions, and tangential remarks. From these counts, clinicians calculate the percentage of CIUs and CIUs per minute to quantify communicative informativeness and efficiency. CIUs are crucial in the aphasia and speech-language pathology literature because they provide a standardized, objective measure of how effectively a person conveys meaningful content in spontaneous language, independent of grammaticality or fluency. This makes the CIU metric uniquely sensitive to real-world communicative ability and valuable for tracking change over time, evaluating treatment outcomes, and comparing discourse performance across individuals and contexts—bridging the gap between linguistic form and communicative function. Computing CIUs, at present, is heavily time consuming and must be done manually by trained specialists.

With this in mind, existing research describes two problems which motivate this work. First, Manir et al. (2024) claimed, "[a]utomatic recognition of aphasic speech is difficult due to various impairments and limited training data" (pg. 1). We consider the problem of limited training data to be critical. Moreover, in this context, *automated recognition* can be understood as a machine learning (ML) implementation. Yet, data availability and data quality are well-known limitations affecting ML too. Consequently, ML researchers broadly have begun turning to *synthetic data*[1] for purposes of training and evaluation. To date, there has not been published discussion regarding the generation of synthetic data as PWA transcripts.

The second existing problem also concerns training data, albeit from a different direction. Likewise well-known are the challenges associated with manual analysis of transcripts in terms of labor effort, error rates, and cost (Stark et al., 2021; Day et al., 2021; Casilio et al., 2023). Such challenges extend to training clinicians to perform the analyses as well (Leaman & Edmonds, 2019; Obermeyer, Leaman & Oleson, 2025). Whilst some corpuses have become available supporting transcript analysis in clinical populations (e.g., the TalkBank project, such as AphasiaBank [MacWhinney et al., 2011]), these datasets are still small relative to the needs of ML systems.

For example, at present, AphasiaBank contains data from ~600 persons with aphasia, ~100 producing data at two test-retest timepoints (Stark et al., 2023; Stark et al., 2025), yet LLMs and other tools are generally trained on billions of tokens (Xue et al., 2023). Thus, in some sense, one can surmise the second problem engenders the first. In other words, because manual analysis is tedious and costly, availability of sufficient data to train robust and reliable ML solutions is inhibited.

Accordingly, the purpose of this work is to describe the construction and preliminary validation of a system to generate reliable person with aphasia (PWA) transcripts based on the common Cat Rescue descriptive task from the AphasiaBank protocol. The Cat Rescue is a single picture derived from a well-established protocol from Nicholas & Brookshire (1993). Generally, single picture descriptions are the most used task to elicit spoken language across adult clinical populations (Bryant et al., 2016), such as the pervasive use of the Cookie Theft single picture description (from the Boston Diagnostic Aphasia Examination; Goodglass, Kaplan & Weintraub, 2001) to evaluate language use in cognitive impairment (e.g., dementia) (Fromm et al., 2024; Giles, Patterson & Hodges, 1996; Berube et al., 2019).

Our purpose is not without grounding. Indeed, synthetic data generation has gradually become mainstream in ML research (Lu et al., 2023). The shift from traditional data sources to synthetic has been driven by insufficient data volume, inferior data quality, and privacy concerns. The latter is rather pertinent to ML applications in healthcare (Hittmeir, 2019; Dankar, 2021) where patient health information is tightly regulated. Moreover, the large language model (LLM) arms race is fueling the incredible thirst for novel data (Goyal & Mahmoud, 2024). Such is taking place at the same time aphasia research (Cong et al., 2024; Kurland et al., 2025) is exploring the practical application of LLMs.

---

[1] Synthetic data are, "...artificially annotated information generated by computer algorithms or simulations" (Lucini, 2021, pg. 11).

**Method**

Given the grounding for this work, the following methods detail how we constructed two systems to generate synthetic PWA transcripts. We selected the Cat Rescue description task as the overarching clinical context (Fig 1). This single picture description task has strong representation in the literature, being part of the AphasiaBank protocol which now includes data from >500 persons with aphasia (MacWhinney et al., 2011), and represents one task from the widely-used "picture description" genre.

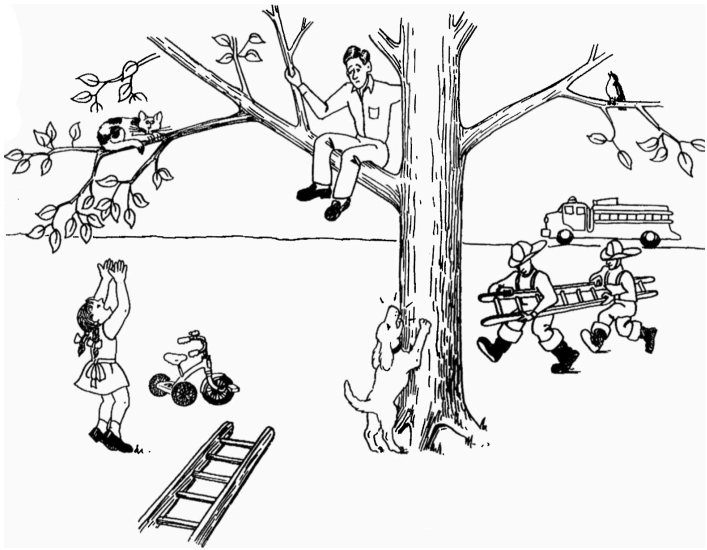

Fig 1. "Cat Rescue" picture, freely available from AphasiaBank ([aphasia.talkbank.org](aphasia.talkbank.org)).

For the first pathway we used procedural programming to modify predefined text with randomly selected elements. The aim here was to construct a lightweight method using fundamental programming techniques. The second pathway employed a set of LLMs, representing a more advanced and modern method.

In general terms, both methods attempt to express a logic described propositionally as follows. The set of words contained in an utterance is defined as

$$\text{Let } W = \{w_1, w_2, w_3, ...\} \quad (1)$$

Then, denote the fixed set of base sentences describing the cat-rescue event as

$$\text{Let } S = \{s1, s2, s3, s4, s5\} \quad (2)$$

Then

$$\text{Let } A = \{a1, a2, a3\} \quad (3)$$

represent the set of augmentation operators corresponding to word dropping, filler insertion, and paraphasia substitution (i.e., {drop,filler,para}).

Finally, we define the set of aphasia severity of Mild, Moderate, Severe, Very Severe classes as

$$\text{Let } \Sigma = \{M, M, S, VS\} \quad (4)$$

Then, for each severity level $\sigma \in \Sigma$ define a mapping as

$$P_\sigma : A \to [0, 1] \quad (5)$$

such that $P_\sigma(a_i)$ gives the probability of applying $a_i$ augmentation to any eligible word $w_j \in W$.

Accordingly, we then define an augmentation function

$$f\sigma : S \to S \quad (6)$$

such that for each sentence $s_k \in S$

$$s'_k = f_\sigma(sk) \quad (7)$$

is produced by stochastically applying the augmentation operators $A$ under the probabilities $P_\sigma$.

Collectively, the operational output becomes The transcript $T_\sigma$ for a given severity class is formed as the ordered concatenation

$$T_\sigma = \langle s'_1, s'_2, s'_3, s'_4, s'_5 \rangle \quad (8)$$

of $T_\sigma$, where each $t_i$ corresponds to a lexical token.

**Procedural Method**
The goal of the procedural method (*procedural_generator.py*, Pittman, 2025) is to generate a large, controlled corpus of synthetic discourse samples modeled on the Cat Rescue picture description task. Generated samples in the corpus leverage severity-specific linguistic characteristics representative of people with aphasia and quantitative CIUs (Nicholas & Brookshire, 1993) annotations suitable for training speech language pathologists (SLPs), clinical researchers, as well as potential ML model fine-tuning and evaluation.

The method included six stages which are described in the next sections. Stages one through three are core synthetic data generation functions whereas stage four and five are utility or helper functions that produce metadata associated with the transcript. After the above, the last stage (i.e., stage six) produces the output in plaintext.

*Input Framework*
The method begins with a configurable set of five base sentences describing the Cat Rescue picture. Base sentences were sampled from existing transcripts (Stark et al., 2023, 2025) and parsed into discrete sentences. The method then applies an aphasia simulation mechanism to get from base sentences to synthetic transcripts.

### *Aphasia Simulation Mechanism*
The procedural pathway includes three probabilistic augmentations designed to simulate speech characteristics associated with aphasia. Individual augmentation values are configurable with values ranging from 0.0 (no augmentation) to 1.0 (full augmentation). For clarity, higher values increase the frequency of omissions, disfluencies, and lexical errors, thereby producing speech that reflects greater impairment severity. This structure operationalizes graded symptom simulation and enables systematic variation of linguistic degradation consistent with clinical patterns of aphasic discourse.

The set of augmentations is encapsulated in severity categories, which are broadly construed from classifications on a standard aphasia battery (Western Aphasia Battery - Revised; Kertesz, 2007): Mild, Moderate, Severe, and Very Severe. In other words, severity level determines the probability of each alteration, allowing progressive degradation of linguistic coherence and completeness going from Mild to Very Severe.

The first augmentation is *word dropping* to represent agrammatism and reduced sentence completeness. Then, *filler insertion* takes place to simulate disfluency and hesitation. Third, the method introduces semantic or phonemic errors paraphasia substitutions. Overall, all augmentation is governed by a configurable set of *protected words*. These are words such as *tree* and are terms the method treats as immutable.

### *Transcript Construction*
The aphasia simulation mechanism leads to construction of a synthetic transcript inclusive of each severity category. More specifically, for each severity category, 2,500 transcripts are generated by applying the probabilistic augmentations to each base sentence. The resulting sentences are concatenated into full picture descriptions forming one description per sample description.

### *CIU Computation*
Each transcript is programmatically tokenized to isolate word-level units. Then, the total word count, number of CIUs, and CIU percentage are calculated following established clinical scoring criteria, excluding fillers, conjunctions, and irrelevant items.This is a blind function however and is not intended to produce accurate, reliable scoring. Instead, these metadata are intended to be used for SLP training and also placeholding for future validation work on ML-based discourse analysis.

### *Dataset Composition and Splitting*
The complete generated dataset contains 10,000 synthetic transcripts, evenly distributed across severities ($n = 2,500$). These data into training (80 percent), validation (10 percent), and test (10 percent) subsets to support ML training, fine-tuning, and performance evaluation. However, the data can be concatenated into any desired file and format for training SLPs.

*Output Format*

Each transcript and its associated metadata are output to a JSONL file with entries as standardized fields for transcript text and metadata (e.g. severity label, word counts, and CIU metrics). Separate JSONL files are produced for dataset splits to ensure reproducibility and compatibility with machine-learning pipelines.

**Machine Learning Method**

We also construct and validate a ML method (*llm_generator.py*, Pittman, 2025) to generate a synthetic corpus of discourse associated with the Cat Rescue description task. For this method, we selected two open source LLMs: Llama 3.1 8b instruct and Mistral 7b instruct. Both models are auto-regressive and come pretrained and fine-tuned for following instructions as prompts.

*Prompt Template Engineering*

After identifying which LLMs to trial, we developed a set of instruction-based prompt templates consisting of *system* and *user* segments (*promptpack*, Pittman, 2025). Each system segment primes the model behavior. User segments then represent a task request and desired criteria for the task. Of note, templates are paired to each respective model and are not cross-validated.

The development process defined ground truth by isolating exemplars from existing transcripts (Stark et al., 2023) exhibiting spoken language impairments from PWA alongside spoken language from cognitively healthy peers, which displayed accurate and reliable Cat Rescue descriptive language. From the isolated exemplars, we identified shared anchors and core facts similar to the immutable words concept employed by the procedural method. Examples here would be *cat up the tree* or *firefighters approaching*.

Table 1

*Prompt template examples of language variety and diversity architecture*

| Severity | Language Architecture |
|---|---|
| Mild | Speak naturally and concisely. Include most key details. Occasional brief hesitations or minor word-finding issues are acceptable. Avoid technical terms |
| Moderate | Use some circumlocutions, a few phonemic/semantic errors, and several hesitations (um/uh/…). Keep sentences short. Mention several key details but omit a couple. |
| Severe | Telegraphic style. Short phrases. Missing function words. Frequent hesitations. Include some errors, 1-2 nonwords, and at least one self-repair [like this] |
| Very Severe | One-two-word bursts, long pauses, frequent failed starts. 1-2 neologisms allowed. May mislabel items |

As well, we follow the same severity architecture as used in the procedural method (i.e., Mild, Moderate, Severe, and Very Severe). Departing from the procedural method, we employ four prompt templates for each severity intending to introduce diversity and variety across generated synthetic transcripts (Table 1). Coupled with the non-determinism inherent to an LLM, the generated transcripts ideally represent diversity of *severe* aphasia.

### *Model Configuration*
Both selected models accept standard hyperparameters[2]. For this study, we explicitly set *temperature* to 0.7, *top_p* to 0.9, and repetition_penalty to 1.0. In simple terms, we allowed the transcript generation to exercise high creativity, use most likely tokens whose probabilities add up to at least 90%, and not penalize repetition at all. Of course, these hyperparameters are configurable in the synthetic data generator.

### *LLM Interactions*
The core of the ML method is using the aforementioned prompt sets to generate synthetic transcripts. To this end, we constructed a LLM pipeline (*llm_generator.py*, Pittman, 2025) to read in the prompts and output four transcripts per severity level. Additionally, the LLM generates metadata for each transcript as outlined in the procedural method.

### *Output Format*
Similar to the procedural method, we constructed the ML method to output in a standard JSONL format. The method also outputs a comma-separated (CSV) file for ease of downstream evaluation.

### **Preliminary Results**
We set out to measure whether the synthetic methods reproduce an expected gradation of linguistic impairment (Mild to Very Severe) across several lexical metrics after constructing the two methods. The results are not intended to convey precision, rather we offer an evaluation of a determinant as to whether additional work for one or both methods may be viable.

We ran both synthetic transcript generators using the same hardware and software environment. Hardware consisted of a M4 Macbook Pro. Software included Python 3.9.6 and standard packages. A sample of both procedural and LLM generated transcripts are shown in Appendix A. Such are a limited but representative snapshot of what the generator methods produce in output.

Generated transcript (*data/*, Pittman, 2025) output from both methods were subjected to identical data analysis procedures. The initial procedure consists of lexical richness calculations (Yang & Zheng, 2024). More specifically, the output consists of Type-Token Ratio (TTR), Number of Different Words[3] (NDW), Lexical Density, Number of Words, and Average Length of Words. The data analysis program (*transcript_analysis.py*, Pittman, 2025) is included for reproducibility and future analysis of additional synthetic transcripts.

---

[2] Temperature governs the randomness of its output. Nucleus sampling (i.e., top_p) is the cutoff for the cumulative probability of tokens considered for the next word. Repetition penalty dictates the probability of tokens that have already appeared appearing again.

[3] We use NDW-ER50 when the transcript sample contains 50 or more *words*. Otherwise, basic NDW is calculated.

*Procedural Method*

The procedural method (Sample A) yields one set of outputs (Table 1) per aphasia severity level based on a total sample size of 1,000 synthetic transcripts. The sample includes 250 transcripts per severity level.

Table 1.
Lexical richness of procedural synthetic data (Sample A)

| TTR | NDW | LD | Words | Avg |
|---|---|---|---|---|
| **Mild** | | | | |
| 0.62 | 22.73 | 0.52 | 36.52 | 3.74 |
| **Moderate** | | | | |
| 0.63 | 21.82 | 0.53 | 34.55 | 3.76 |
| **Severe** | | | | |
| 0.64 | 20.58 | 0.54 | 31.99 | 3.78 |
| **Very Severe** | | | | |
| 0.65 | 19.51 | 0.55 | 29.96 | 3.8 |

*Machine Learning Method*

The ML (i.e., LLM) method yields two outputs (Table 2 and Table 3) per severity level. Each output consisted of 32 synthetic transcripts with eight per severity level. First, we ran the Mistral 7b Instruct prompts (Sample B) against a local model. This yields the following lexical richness description.

Table 2.
Lexical richness of Mistral synthetic data (Sample B)

| TTR | NDW | LD | Words | Avg |
|---|---|---|---|---|
| **Mild** | | | | |
| 0.67 | 38.02 | 0.52 | 109.75 | 3.82 |
| **Moderate** | | | | |
| 0.69 | 39.41 | 0.48 | 123.38 | 3.57 |
| **Severe** | | | | |
| 0.72 | 38.59 | 0.56 | 90.38 | 3.62 |
| **Very Severe** | | | | |
| 0.77 | 27.5 | 0.55 | 35.38 | 3.72 |

Next, we ran the Llama3.1-8b Instruct prompts (Sample C) against a local model which produced the following lexical richness description.

Table 3.
Lexical richness of Llama synthetic data (Sample C)

| TTR | NDW | LD | Words | Avg |
|---|---|---|---|---|
| **Mild** | | | | |
| 0.61 | 38.51 | 0.49 | 156.75 | 3.8 |
| **Moderate** | | | | |
| 0.77 | 43.7 | 0.52 | 144.5 | 3.93 |
| **Severe** | | | | |
| 0.82 | 44.94 | 0.58 | 116.75 | 3.83 |
| **Very Severe** | | | | |
| 0.94 | 47.93 | 0.74 | 99.25 | 4.29 |

### Word and CIU Analysis

For completeness, we analyze (*augment_llm_metrics.py*, Pittman, 2025) the frequency of Words and CIUs contained in the generated synthetic transcripts. We offer two views: (a) broad analysis across all generated transcripts by method; and (b) a per severity breakdown by method. With that stated, we stress again that CIU identification is not clinically reliable in this work. Neither method infers context. Word frequency data are reliable as such do not rely on inferences or relations. That being true, across severity levels, the procedural method yields consistently high CIU density and stable word counts whereas the LLM samples display different lexical profiles and CIU rates as captured in the metrics. The *average of means* (Table 4) provides a single comparative snapshot.

Table 4.
*Average of Means for word and CIUs across synthetic data generation methods*

| Method | Word Count | Avg Word Count | CIUs | % CIUs |
|---|---|---|---|---|
| Sample C | 117.47 | 22.61 | 106.59 | 91.09 |
| Sample B | 72.84 | 10.53 | 64.69 | 90.39 |
| Sample A | 34.99 | 7 | 28.26 | 81.1 |

*Note: Sample C is Llama 3.1 8b Instruct, Sample B is Mistral 7b Instruct, and Sample A is procedural.*

Then, the per-severity description (Table 5) reveals telltale declines in CIU percentage with increasing severity for the procedural data and the corresponding patterns for LLM outputs.

Table 5.
*Average of Means for word and CIUs across methods per severity level*

| Method | Severity | Total Word Count | Avg Word Count | CIUs | % CIUs |
|---|---|---|---|---|---|
| Sample C | Mild | 136.62 | 13.24 | 120.75 | 88.65 |
| Sample C | Moderate | 131.12 | 20.1 | 117.62 | 89.59 |
| Sample C | Severe | 106.25 | 23.19 | 97.62 | 91.78 |
| Sample C | Very severe | 95.88 | 33.92 | 90.38 | 94.34 |
| Sample B | Mild | 94.25 | 13.24 | 82.12 | 87.53 |
| Sample B | Moderate | 98.5 | 10.91 | 84.75 | 85.64 |
| Sample B | Severe | 71.75 | 9.63 | 66.38 | 94 |
| Sample B | Very severe | 26.88 | 8.32 | 25.5 | 94.39 |
| Sample A | Mild | 34.62 | 6.92 | 31.52 | 91.26 |
| Sample A | Moderate | 35.4 | 7.08 | 29.55 | 83.81 |
| Sample A | Severe | 34.89 | 6.98 | 26.99 | 77.76 |
| Sample A | Very severe | 35.05 | 7.01 | 24.96 | 71.56 |

*Note: Sample C is Llama 3.1 8b Instruct, Sample B is Mistral 7b Instruct, and Sample A is procedural.*

### Post Hoc Comparison

Finally, we offer a post hoc comparison that emphasizes between-dataset lexical richness fidelity. In other words, how well the synthetic language *on average* resembles human-elicited aphasic speech. To do this, we collect a sample of 12 clinical transcripts (four from known healthy individuals, eight from individuals with some level of aphasia) and analyze those as a baseline. Of note, the sample is secondary data in the context of this study having been previously collected by one of us (Stark, et al., 2025). No PII or PHI is included.

Then, we calculate an average of means for all severities in the procedural (Synthetic A) and ML (Synthetic B and C, respectively) method preliminary results. The resulting composite expresses the *average lexical richness* produced by each generator overall, irrespective of internal severity gradation.

Table 6.
Lexical richness of human-elicited transcripts

| TTR | NDW | LD | Words | Avg |
|---|---|---|---|---|
| **Healthy** | | | | |
| 0.52 | 37.53 | 0.58 | 220.75 | 3.83 |
| **Aphasic** | | | | |
| 0.56 | 29.32 | 0.54 | 76.38 | 3.58 |

Table 7.
Average of means for synthetic transcript methods compared to human transcripts

| Measure | Aphasic (Actual) | Synthetic A (≈mean) | Synthetic B (≈mean) | Synthetic C (≈mean) |
|---|---|---|---|---|
| TTR | 0.56 | ~0.64 | ~0.71 | ~0.79 |
| NDW | 29.32 | ~21 | ~36 | ~44 |
| LD | 0.54 | ~0.54 | ~0.53 | ~0.58 |
| Words | 76.38 | ~33 | ~90 | ~129 |
| Avg | 3.58 | ~3.77 | ~3.68 | ~3.96 |

**Conclusions**

The purpose of this work was to present two methods for the generation of synthetic transcripts associated with the Cat Rescue description task commonly used in assessment of aphasic spoken discourse. One method consisted of procedurally modifying base sentences using simulated aphasia mechanisms. The other method used engineered prompts to generate transcripts from two open source LLMs. With both the goal is to have human-like synthetic transcripts that can be used for SLP training as well as training downstream AI systems for clinical use. With that being said, we offer the following conclusions based on preliminary results and post hoc comparison.

Principally, both methods are operational. Meaning, we are able to generate synthetic transcripts. Secondarily, the methods do differ in lexical richness between each other.

The Synthetic A data successfully mirrors the directional trend for productivity (NDW, total words) but diverges in lexical diversity and compositional balance (TTR, LD, word length). This indicates partial ecological validity, sufficient for structural modeling but requiring further calibration for semantic–lexical realism. Meanwhile, Synthetic B demonstrates moderate ecological fidelity, capturing the correct direction from mild severity through to very severe for NDW, word count, and word length while keeping values within plausible human ranges.

Synthetic C departs from clinical patterns across nearly all dimensions except output length, indicating it over-generates lexical diversity and density inconsistent with aphasic language.

Compared to human transcripts, we suggest Synthetic B captures key aspects of linguistic degradation observed in aphasia, showing realistic directional changes in NDW, word count, and word length. Synthetic C appears to diverge sharply, overrepresenting lexical diversity and complexity. In sum, these findings substantiate the face validity of Synthetic B and suggest targeted constraints to enhance the ecological realism of future synthetic datasets.

### *Limitations*
At face value, an overarching limitation is both transcript generation methods are constrained to a Cat Rescue picture description task. Moreover, SLP interpretation of the synthetic transcripts is confined to a small sample size ($n = 2$). Thus, the preliminary results are suggestive rather than conclusive.

The procedural method has two further or specific limitations. The general procedural algorithm relies on sentence level templates. While configurable, the templates used in this study represent a narrow band of sentences free from aphasic elements. Further, filler and protected words are hardcoded. As such, the method may not include the most common terms and the existing sets are small. Consequently, there is little diversity across synthetic transcripts, particularly within the same severity category.

The ML method has two additional limitations. Foremost, we included only two models with comparatively low parameter sizes. Whereas the largest model in this study has 8 billion parameters, frontier *dense* models at the time of the work have up to 405 billions parameters. As well, trials were run using a small set of engineered prompts with static hyperparameter configurations. Small parameter sizes have lower fidelity in generated text while the lack of ablation study limits the generalizability of the preliminary results.

### *Recommendations*
Based on the preliminary results, we suggest it is possible to generate robust synthetic transcripts across a range of aphasia severities. Such has significance for both training of SLP and potential applications of AI in clinical environments.

Certainly the procedural method requires more work to be viable compared to the ML methods, especially Mistral 7b Instruct (Sample B). Here, we recommend consideration be given to normalizing TTR for sample length to mitigate potential word (i.e., token) count effects. Further, we recommend frequency-based lexical weighting and enhanced discourse lengths.

With the ML method, we recommend consideration be given to the prompt templates. In particular, Llama 3.1 8b might benefit from attention to the prompt modification related to discourse length in a ratio with constrained lexical access. Both models may demonstrate increased performance if lexical density were to be constrained.

*Future work*

Foremost, future work related to practical applications of ML to aphasia clinical tasks (e.g., discourse analysis) ought to consider augmenting ML model training or fine-tuning with synthetic data rather than relying on AphasiaBank alone. Doing so would diversify model applicability and reduce extant biases. Here, we envision establishing a hybrid dataset consisting of various human data, synthetic data training and evaluation splits (e.g., 80/20, 70/30) followed by robust ML model performance assessment.

Extending the ML method to generate transcripts for additional discourse tasks such as narrative storytelling (e.g., Cinderella) represents compelling future work. On one hand, doing so might address the enhanced discourse length recommendation. On the other hand, longer form narratives could introduce confounding variables not present in the shorter description tasks. Thus, both the same LLMs as well as additional models may need to be tested.

Furthermore, as much as existing literature (Dankar & Ibrahim, 2021; Lucini, 2021) suggest synthetic data obviates privacy concerns, we suggest future work investigate to what extent it might be possible to execute model inversion attacks against ML systems applied to clinical aphasia tasks. Model inversion attacks allow an attacker to leverage the output of a trained ML model to reconstruct or infer sensitive information about the original training data.

We also suggest future work to expand synthetic data generation to other tasks such as Narrative (Cinderella) and Procedural (make a PBJ sandwich) is necessary. Doing so would directly address the limitation of this work insofar as the synthetic data generation methods are tightly coupled to Cat Rescue picture descriptions. Similarly, it might be beneficial to test synthetic transcript generation for languages with poor, or no, representation in existing indices (e.g., AphasiaBank) because there are scant samples available.

# Appendix A
# Examples of Procedural and LLM Generated Transcripts

**Procedural Examples**

Example 1:
The um cat is stuck yup up tree. The little girl called her father to use a ladder. But the ladder fell uh-huh. They called the fire department nope. The fire department came to rescue so the uh cat.

Example 2:
The cat and is up the tree. little girl called oh her father to a ah ladder. But ladder. They uh called the ah fire department. The fire yup department came to rescue oh the cat.

Example 3:
Cat uh-huh is stuck up the tree nope. The so little girl so called uh-huh mother to and use ladder. But ladder. They yup called oh fire department so. The oh fire department came to the uh-huh cat oh.

Example 4:
The then cat and up tree. The um-hum boy called father to ladder. But the yup ladder so fell. They um-hum called fire department then. The fire department uh-huh to rescue the dog.

**LLM Examples**

*Llama 3.1 8B*

Example 1:
It's a cat rescue scene. There's a woman kneeling down next to a small, old-looking house with a broken roof. She's holding a little kitten in her arms. The kitten looks scared but seems okay. It's trying to hide its face behind the woman's hand.

Example 2:
Um, there's like this big tree... and it has branches that have broken off from a storm I think. And then you see one of these animals, uh, it was supposed to be a dog, no wait, it's actually a cat stuck up high in the branch.

Okay so some people come running over because they heard all the commotion and they're trying to figure out how to get the cat down safely without hurting it. One person starts looking around for something sturdy enough to climb up with while another person tries to distract the cat by calling its name softly.

They manage to find a long rope which they tie together with some other stuff to create kind of a makeshift pulley system and carefully lower the cat down onto a blanket someone else brought over. Everyone breathes a sigh of relief when the cat lands safe and sound on the ground.

Example 3:
A cat was trapped in an attic above a house with no visible entry point except through a very small window.

[rep] hole

The woman called her neighbor to help and they were both able to climb up to where she could access it safely.

They used some tools from home like a screwdriver and pliers to widen the opening further so that the animal wouldn't be injured when exiting.

It took them about twenty minutes before finally managing to get her out safely

Example 4:
It's like you have these cats and umm they're so soft I mean fluffy yeah it smells good too *pauses* there was this one little ball of fur in my arms all squirmy like no wait other cat okay then oh kitty kitten cute tiny whiskers ears... thingy feline rescue place taking care give food water sleeping space quiet spot warm cozy blanket cuddle purr meow mew loud m-e-e-om *whispers softly*.

**Mistral**
Example 1:
Alright, so we have a cat stuck in a tree, it's on the left side about halfway up. There's a firefighter climbing a tall ladder towards the cat. Another firefighter is holding the base of the ladder steady. The cat seems scared but not injured. I can hear sirens approaching, more help might be coming soon.

Example 2:
Alright, so we've got this situation where there's a dog stuck up in a tall oak tree. A group of folks are gathered below, discussing plans to save the animal. One person suggests using a long ladder while another proposes calling an animal rescue service. They decide to do both simultaneously, setting up the ladder cautiously while waiting for help to arrive. The trapped dog barks anxiously, occasionally meowing due to confusion, as it clings tightly to the tree branch.

Example 3:
Uh, so there's this... umm... cat, see? It's stuck in a tall tree, y'know? I mean, it climbed up high and now it can't get down. And there's, like, no one around to help, right? So, I was thinking maybe we could send someone up, you know, professional person, maybe? They got tools, right? Tools that can reach him. Or, uh, I don't know, maybe call for some sort of rescue vehicle? Something big enough to grab hold of the tree and pull it out from under the cat, you follow me? Yeah, that should work, yeah. Help is on its way, yes, definitely. Just need to make sure it gets here quick before our feline friend gets too scared or something. Hopefully, everything will be alright. Fingers crossed!

Example 4:

Uh... found... cat, umm, in tree, uhh.. very scared, needs help, food and water... probably hurt, need vet checkup... call animal shelter, find good home.